\documentclass[10pt,conference,a4paper]{IEEEtran}
\pdfoutput=1
\usepackage{amsmath}

\usepackage{float}
\usepackage{dblfloatfix}

\usepackage{graphicx}
\usepackage{multirow}
\usepackage{cases}

\ifCLASSINFOpdf
\else
\fi

\begin{document}
%
\title{Not 3D Re-ID: Simple Single Stream 2D Convolution for Robust Video Re-identification }

\author{\IEEEauthorblockN{Toby P. Breckon}
\IEEEauthorblockA{Department of Computer Science\\
Department of Engineering\\
Durham University,UK\\}
\and
\IEEEauthorblockN{Aishah Alsehaim}
\IEEEauthorblockA{Department of Computer Science\\
Durham University, UK\\
\\
Department of Computer Science\\
Princess Nourah Bint Abdulrahman University, SA}
}


%


\maketitle

\begin{abstract}
Video-based person re-identification has received increasing attention recently, as it plays an important role within surveillance video analysis. Video-based Re-ID is an expansion of earlier image-based re-identification methods by learning features from a video via multiple image frames for each person. Most contemporary video Re-ID methods utilise complex CNN-based network architectures using 3D convolution or multi-branch networks to extract spatial-temporal video features. By contrast, in this paper, we illustrate superior performance from a simple single stream 2D convolution network leveraging the ResNet50-IBN architecture to extract frame-level features followed by temporal attention for clip level features. These clip level features can be generalised to extract video level features by averaging without any significant additional cost.  Our approach uses best video Re-ID practice and transfer learning between datasets to outperform existing state-of-the-art approaches on the MARS, PRID2011 and iLIDS-VID datasets with $89.62 \%$, $97.75\%$, $97.33\%$ rank-1 accuracy respectively and with $84.61\%$ mAP for MARS, without reliance on complex and memory intensive 3D convolutions or multi-stream networks architectures as found in other contemporary work. Conversely, our work shows that global features extracted by the 2D convolution network are a sufficient representation for robust state of the art video Re-ID.
\end{abstract}


%
\IEEEpeerreviewmaketitle

\section{Introduction}
Video based person re-identification is a popular research area within computer vision gaining increasing attention due to a wide range of potential applications, such as intelligent video surveillance and automated security systems.  In addition to its promising application, the emergence of deep learning techniques and the availability of large-scale video surveillance datasets  \cite{zhengMARSVideoBenchmark2016b},\cite{hirzerPersonReidentificationDescriptive2011b},\cite{wangPersonReidentificationSystem2014d} has resulted in the rapid evolution of this task domain within recent years.\newline
Video person re-identification refers to the task of matching a person in a query surveillance video, to the same person within other videos from multiple non-overlapping cameras. However, it is a very challenging problem due to the large variations of human pose, occlusion, differing camera viewpoints, illumination and background scene clutter.\newline 
Extensive research has been proposed to improve the video re-identification task \cite{lengSurveyOpenWorldPerson2019}. Video-based Re-ID benefits from the rich multi-frame, temporal information within video to address the task of cross video instance matching. Video-based Re-ID has also benefited from the significant development of deep learning methods by building differing structural approaches that learn discriminative and robust deep features of human subjects in a video \cite{liMultiScale3DConvolution2019},\cite{liGlobalLocalTemporalRepresentations2019a},\cite{houVRSTCOcclusionFreeVideo2019}. However, the most recent state of the art work in the field use combinations of complex 3D convolutions \cite{liMultiScale3DConvolution2019} or multi-stream networks \cite{chenVideoPersonReidentification2018b},\cite{mclaughlinRecurrentConvolutionalNetwork2016b},\cite{xuJointlyAttentiveSpatialtemporal2017} to accomplish this task, both of which are computationally expensive and memory intensive.\newline
Recently, emergent techniques using differing loss functions can be used to generate a robust Re-ID model via the use of CNN \cite{szegedyRethinkingInceptionArchitecture2016a},\cite{wangRankedListLoss2019b},\cite{pathakVideoPersonReID2019a}. In addition, different types of data augmentation techniques, such as Random Erasing Augmentation (REA) \cite{zhongRandomErasingData2017a}, that support the model robustness to partial occlusion, lead to significant improvements in Re-ID accuracy for video Re-ID without the need for complex networks architectures. \newline
The requirement for a simple and effective, yet robust architecture for video person Re-ID is a key challenge that looks to enable real-time video analysis within a viable computational bound and memory footprint. However, currently many leading approaches rely on complex 3D convolution \cite{liMultiScale3DConvolution2019} and multiple stream architectures  \cite{chenVideoPersonReidentification2018b},\cite{mclaughlinRecurrentConvolutionalNetwork2016b},\cite{xuJointlyAttentiveSpatialtemporal2017}.\newline
By contrast in this paper, we differentiate ourselves from prior work in the field \cite{liMultiScale3DConvolution2019},\cite{liGlobalLocalTemporalRepresentations2019a},\cite{chenVideoPersonReidentification2018b},\cite{mclaughlinRecurrentConvolutionalNetwork2016b},\cite{xuJointlyAttentiveSpatialtemporal2017} by using only 2D convolutional operations and additionally a single stream network architecture. Our 2D single stream architecture uses ResNet50-IBN \cite{panTwoOnceEnhancing2018b} as the frame feature extractor followed by a simple temporal attention layer to aggregate clip level features \cite{gaoRevisitingTemporalModeling2018a}. An illustration of our proposed method is shown in Figure \ref{Figure 1}.\newline
In addition, we employ multiple loss functions Label Smoothing (LS) loss \cite{szegedyRethinkingInceptionArchitecture2016a}, RLL loss \cite{wangRankedListLoss2019b}, center loss \cite{wenDiscriminativeFeatureLearning2016a}, erasing-attention loss \cite{pathakVideoPersonReID2019a} and use well established training strategies such as Batch Normalisation Neck (BNNeck) \cite{luoStrongBaselineBatch2019}, Random Erasing Augmentation (REA) \cite{zhongRandomErasingData2017a} and Warmup \cite{fanSphereReIDDeepHypersphere2019a}, to boost the performance of our resulting model. Finally, we leverage a range of best practices in video ID in terms of network training, frame window selection, training batch size and transfer learning \cite{lengSurveyOpenWorldPerson2019}.\newline
The main contributions of this paper are:

\begin{itemize}
\item we propose an efficient novel single stream architecture based on 2D convolution operations, capable of achieving state of the art performance on MARS \cite{zhengMARSVideoBenchmark2016b}, PRID2011 \cite{hirzerPersonReidentificationDescriptive2011b} and iLIDS-VID \cite{wangPersonReidentificationSystem2014d} outperforming prior work of \cite{liMultiScale3DConvolution2019},\cite{chenVideoPersonReidentification2018b},\cite{mclaughlinRecurrentConvolutionalNetwork2016b} which are conversely reliant upon both computational expensive 3D convolutions \cite{liMultiScale3DConvolution2019} and multiple stream architectures \cite{chenVideoPersonReidentification2018b},\cite{mclaughlinRecurrentConvolutionalNetwork2016b},\cite{xuJointlyAttentiveSpatialtemporal2017}.

\item we illustrate that the performance of such an architecture can be readily enabled following the combined best practice approaches of Batch Normalisation Neck (BNNeck) \cite{luoStrongBaselineBatch2019}, Random Erasing Augmentation (REA) \cite{zhongRandomErasingData2017a} and Warmup \cite{fanSphereReIDDeepHypersphere2019a} to jointly boost the performance of the model.

\item we report state of the art performance on the MARS ($89.62\%$), PRID2011 ($97.75\%$) and iLIDS-VID ($97.33\%$) for rank-1 accuracy. 
\end{itemize}

\section{Related work}

There are two branches for person Re-ID within contemporary work enabled by deep learning image based Re-ID \cite{wangPersonReidentificationSystem2014d},\cite{sunPartModelsPerson2018d},\cite{suPoseDrivenDeepConvolutional2017a},\cite{pedagadiLocalFisherDiscriminant2013a} and video based Re-ID \cite{pathakVideoPersonReID2019a},\cite{zhaoAttributedrivenFeatureDisentangling2019},\cite{houVRSTCOcclusionFreeVideo2019},\cite{subramaniamCoSegmentationInspiredAttention2019a},\cite{guTemporalKnowledgePropagation2019a},\cite{liGlobalLocalTemporalRepresentations2019a}. This section briefly reviews the relevant prior work in video based Re-ID that form the basis for our research within this paper.\newline 
 Reliable feature representations for person Re-ID in current research are extracted by tailor-made based architectures \cite{sunPartModelsPerson2018d},\cite{suhPartAlignedBilinearRepresentations2018a},\cite{wangPersonReidentificationCascaded2018a} or generic convolutional neural network (CNN) architectures \cite{pathakVideoPersonReID2019a},\cite{zhaoAttributedrivenFeatureDisentangling2019},\cite{liGlobalLocalTemporalRepresentations2019a}. Tailor-made architectures are designed to consider the structure of the human body to reduce the effect of occlusion and to alleviate false detections. Most recent researche uses generic CNN architectures as a feature extraction network such as ResNet-50 \cite{pathakVideoPersonReID2019a},\cite{liGlobalLocalTemporalRepresentations2019a} and ResNet-18 \cite{zhaoAttributedrivenFeatureDisentangling2019}. \newline
In addition to spatial feature extraction, temporal information is a significant component of contemporary video based Re-ID. Recent researches use differing ways to aggregate such temporal features with average or max pooling \cite{chenVideoPersonReidentification2018b},\cite{gaoRevisitingTemporalModeling2018a},\cite{liuQualityAwareNetwork2017} and attention models \cite{zhouSeeForestTrees2017b},\cite{sunPartModelsPerson2018d},\cite{liDiversityRegularizedSpatiotemporal2018a} common strategies to aggregate such video frame features. The attention model aggregation focuses on selecting the most informative frames, while average and max pooling treats all the frames equally. Whilst, many techniques use optic flow \cite{chenVideoPersonReidentification2018b},\cite{mclaughlinRecurrentConvolutionalNetwork2016b},\cite{xuJointlyAttentiveSpatialtemporal2017}, many person Re-ID applications require real-time performance precluding the use of such computationally expensive techniques.  Alternatively, recurrent CNN are also explored to capture the temporal structure and aggregate temporal features within videos \cite{chenVideoPersonReidentification2018b},\cite{xuJointlyAttentiveSpatialtemporal2017},\cite{zhouSeeForestTrees2017b}.\newline 
More recently 3D convolution has been adopted for video feature learning in video person Re-ID, as it directly extracts spatial-temporal features \cite{qiuLearningSpatioTemporalRepresentation2017a},\cite{tranLearningSpatiotemporalFeatures2015a}. Multi-scale 3D (M3D) CNN \cite{liMultiScale3DConvolution2019} uses 3D convolutions to extract spatial-temporal features but requires a significantly larger number of parameters to be optimised resulting in both additional computational complexity and an increased memory footprint for both training and inference.\newline
In this paper, in contrast to this prior work via 3D convolution \cite{liMultiScale3DConvolution2019} and multi-stream architectures \cite{chenVideoPersonReidentification2018b},\cite{mclaughlinRecurrentConvolutionalNetwork2016b},\cite{xuJointlyAttentiveSpatialtemporal2017}, we will demonstrate that simple 2D convolution using well defined training strategies across a single stream architecture can exceed prior state of art performance without the additional overhead of such complexities.\newline
\section{Method}
\begin{figure*}
  \renewcommand{\figurename}{Figure}
  \includegraphics[width=\linewidth]{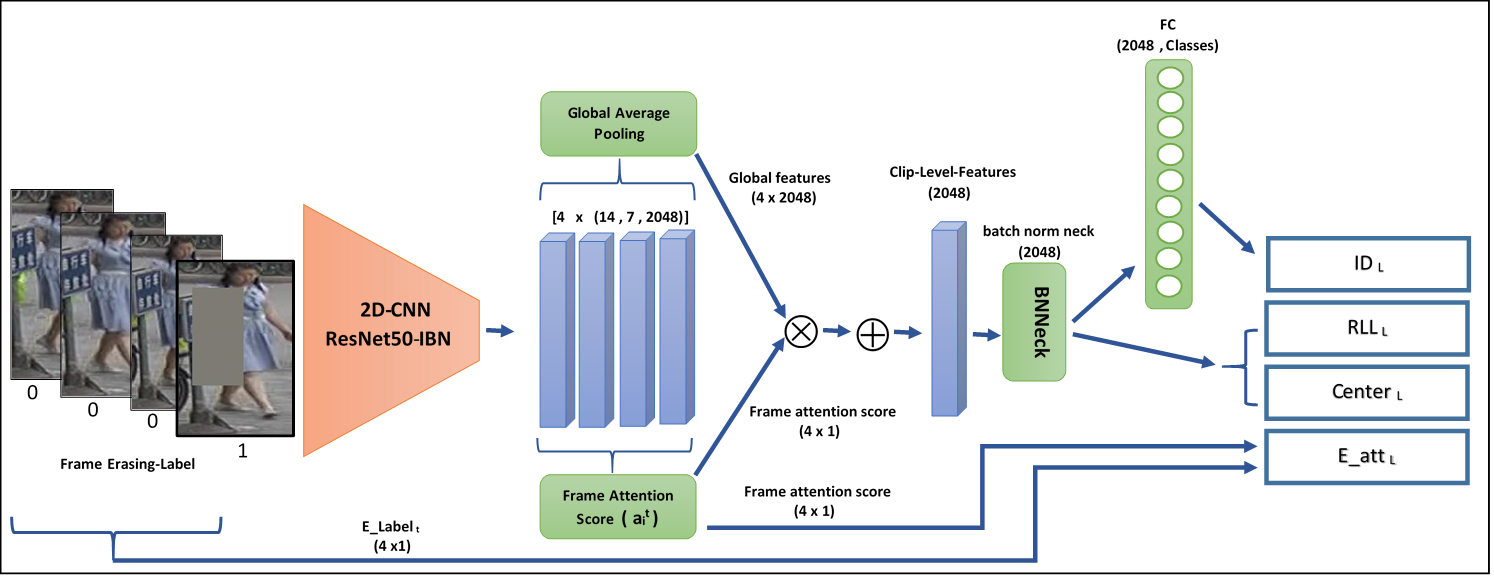}
  \caption{The proposed architecture: global features extracted by the ResNet-50-IBN-a \cite{panTwoOnceEnhancing2018b} followed by simple spatial-temporal attention \cite{gaoRevisitingTemporalModeling2018a} ($\otimes$ indicates pairwise multiplication; $\oplus$ indicates summation). }
  \label{Figure 1}
\end{figure*}

We will introduce the simple yet effective Re-ID model, using spatial-temporal aggregation within a clip that can be generalised to video level aggregation without any significant additional costs. The model is trained using multiple loss functions to find the best representation for a clip containing a target human subject. Our evaluation illustrates that the model effectively generalises to extract effective video level features while minimising the required computational effort for training via the avoidance of the need for explicit 3D convolution within the proposed architecture.\newline
The model learns to produce clip level features, $C$, at the training stage by choosing random frames, $T$, from the tracklet. At inference time, all of the frames in a tracklet are used to produce the video level feature by dividing the tracklet to a number of clips as $P_{vid}$ $( C_1, C_2, ..., C_m)$, each clip has $T$ frames, where $T$ is the number of selected frames to train the network.\newline
Our 2D CNN architecture (Figure \ref{Figure 1}), extracts frame level features, these features are then aggregated using simple temporal attention layers to represent clip-level features. At inference time, the clip-level features are then fused by taking the average of all the clip-level features to represent the gallery and query videos. The overall architecture of our proposed method is shown in Figure 
\ref{Figure 1}.
\subsection{Video Spatial-Temporal Features}
To generate spatial-temporal features, we adopt the identified CNN architecture (ResNet50-IBN-a \cite{panTwoOnceEnhancing2018b}) to extract video images features. After extracting the spatial feature of each image, we use the output of the global average pooling layer of the CNN as the embedding vector for each frame. Subsequently, the temporal feature needs to be aggregated through all the frames.
The results of the previous comparison study \cite{gaoRevisitingTemporalModeling2018a}, shows that temporal attention is the most efficient way to capture temporal information among the sequence of frames in the video as compared to average/max pooling and recurrent neural network (RNN) aggregation. Temporal attention is preformed to obtain attention score ${a_i^t}$ for each frame $f_i^t$ in clip $C_i$ where $ t \in [1, T]$. The frame feature $f_i^t$ of a clip $C_i$ are weighted and averaged to represent clip level features as follows:
\begin{equation} \label{eu_eqn}
F_{C_i} =\frac{1}{T} \sum_{t=1}^{T} a_i^t  \;\; f_i^t\\
\end{equation}
The spatial-temporal attention is preformed using 2D convolution with an input dimensionality of $2048$, the output of the CNN, and the resulting outputs are dimensionality of
 256 following the settings of \cite{gaoRevisitingTemporalModeling2018a}. This spatial attention followed by temporal 1D convolution on the frame-level features generates temporal attentions $s_i^t$. The final frame attention score ${a_i^t}$ is calculated using $softmax()$ \cite{zhouSeeForestTrees2017b}. 
\subsection{Training Strategies}\label{sub:Tr}
The need for a more discriminative video feature representation is ever present within Re-ID in order to distinguish between different people under challenging conditions. Adding a batch normalization (BN) layer, BNNeck \cite{luoStrongBaselineBatch2019}, to be applied to features extracted by the 2D CNN and before the deep metric learning (DML) loss functions, such as center loss \cite{wenDiscriminativeFeatureLearning2016a}, triplet loss \cite{schroffFaceNetUnifiedEmbedding2015a} or rank list loss \cite{wangRankedListLoss2019b}, result in  more discriminative features. In addition to adding a BNNeck layer, the bias of the fully-connected layer ($FC$) needs to be removed and initialised by applying Kaiming initialisation proposed in \cite{he2015delving}. The authors of \cite{luoStrongBaselineBatch2019} suggest using the features of the clip before the BNNeck layer as an input to the classifier, $FC$ layer, but our experiments conversely show that adding BNNeck layer before both of the $FC$ layer and DML loss functions results in more robust features, which boost the learning scheme of the Re-ID model. This finding was also supported by the authors of \cite{pathakVideoPersonReID2019a}.\newline
One of the key issues for Re-ID is dealing with occlusions, wherein, some frames which are part of the subject are occluded. Recent research \cite{houVRSTCOcclusionFreeVideo2019} proposes a new model using a generative adversial network (GAN) to deal with the occlusions issue in Re-ID videos. However, the use of Random Erasing Augmentation (REA) \cite{zhongRandomErasingData2017a}, can force the resulting model to deal with this issue without extra overhead of an additional GAN architecture network to generate the occluded part of the frames. REA randomly erases a rectangular region within the training imagery during learning. This type of augmentation deals with the partial occlusion and improve the generalisation ability of Re-ID model.\newline
Using a variable learning rate has been shown to have a significant impact on CNN model performance \cite{fanSphereReIDDeepHypersphere2019a},\cite{luoStrongBaselineBatch2019}. The authors in \cite{luoStrongBaselineBatch2019}, support the warmup strategy proposed by \cite{fanSphereReIDDeepHypersphere2019a} to bootstrap the network for superior performance. Using multiple learning rates, starting from $3.5 \times 10^{-5}$ followed by a linearly increase in the first 10 epochs until reaching $3.5\times10^{-4}$. Subsequently, the learning rate at epoch 40 is decayed back to $3.5\times10^{-5}$. Another decay to the learning rate, is applied at epoch 70 to $3.5 \times 10^{-6}$ and it remains at this level until the end of the training phase. Subsequently, the learning rate, $LR$, at epoch $E$ is computed as follows: 
\begin{equation}
LR(E)=\left\{\begin{array}{ll}
{3.5 \times 10^{-5} \times \frac{E}{10} }  & {\text { if } E \leq 10} \\
{3.5 \times 10^{-4}} & {\text { if } 10< E \leq 40} \\
{3.5 \times 10^{-5}}  & {\text { if } 40< E \leq 70} \\
{3.5 \times 10^{-6}} & {\text { if } 70< E \leq 120}\\
\end{array}\right.
\end{equation}
The importance of transfer learning cannot be ignored, if we can prior training for a related problem task \cite{lengSurveyOpenWorldPerson2019}. Here, we investigate the use of transfer learning our Re-ID model using the MARS dataset \cite{zhengMARSVideoBenchmark2016b}, the largest and most challenging video Re-ID datasets presently available, in order to initialise the model parameters. For this purpose, we compare the results of  PRID2011 \cite{hirzerPersonReidentificationDescriptive2011b} and iLIDS-VID \cite{wangPersonReidentificationSystem2014d} in two cases. In the first case, when the CNN is pre-trained on the ImageNet \cite{he2015delving} dataset, whiles in the second case, when the CNN is pre-trained on MARS \cite{zhengMARSVideoBenchmark2016b} dataset (Table \ref{table:4}). We have also examined the results of fine tuning our model by a sequence of training, pre-training on MARS \cite{zhengMARSVideoBenchmark2016b} followed by training on iLIDS-VID \cite{wangPersonReidentificationSystem2014d} dataset, (Section \ref{sec:ev}).
\subsection{Multi-loss Function}
The optimisation of the CNN model weights is guided by the use of differing loss functions to produce best spatial-temporal representation of the clip.\newline
Most recent proposed Re-ID approaches \cite{pathakVideoPersonReID2019a},\cite{luoStrongBaselineBatch2019} use a modified version of the cross-entropy loss to prevent overfitting, Label Smoothing (LS) \cite{szegedyRethinkingInceptionArchitecture2016a}, as it encourages the model to be less confident on the training set in order to enforce generalisation to unseen examples. This loss is usually called identification (ID) loss, as it calculates the loss of predicted ID to the truth labels, defined as:
\begin{equation} 
ID_{L}= \sum_{i=1}^{N} - q_i \log (pre_i).
\end{equation}
where $N$ is the number of person subjects. Given a video of person $i$, $pre_i$ is the ID prediction likelihood of class $i$ and $q_i$ is constructed as follows:
\begin{equation}
   q_i=
   \begin{cases}

  1 - \frac{N-1}{N} \epsilon , & \text{if}\  i=y\\
  \frac{\epsilon}{N}, & \text{otherwise}\
  \end{cases}
\end{equation}
 where $y$ as the ground-truth ID label and $\epsilon$ is used to encourage the model to be less confident on the training set. In our experiments, $\epsilon= 0.1$. \\
 In addition, we make use of deep metric learning (DML) that uses the embedding extracted by the model to learn semantic similarity information among data points, which can boost the model learning, such as center loss \cite{wenDiscriminativeFeatureLearning2016a}, triplet loss \cite{schroffFaceNetUnifiedEmbedding2015a} or rank list loss \cite{wangRankedListLoss2019b}. Our model is trained using ranked list loss (RLL) \cite{wangRankedListLoss2019b} and center loss \cite{wenDiscriminativeFeatureLearning2016a} as embedding loss functions.\\
 In our approach, we replace the triplet loss \cite{schroffFaceNetUnifiedEmbedding2015a}, used to train most recent proposed Re-ID models \cite{subramaniamCoSegmentationInspiredAttention2019a}, \cite{luoStrongBaselineBatch2019}, with the ranked list loss (RLL) \cite{wangRankedListLoss2019b} that learns a hypersphere for each class in addition to forcing distance between a positive Re-ID pairing to be smaller than a constant margin. The RLL \cite{wangRankedListLoss2019b} loss, by learning the hypersphere for each class, will avoid intra-class data distribution that may occur in other loss functions, such as triplet loss \cite{schroffFaceNetUnifiedEmbedding2015a}. RLL \cite{wangRankedListLoss2019b} is used to force a distance between negative samples to be greater than specific threshold $\alpha$, whilst positive samples are pulled closer than a threshold of $\alpha - m$, where $m$ is the margin. The pair-wise constraint given an image ${x}_{i}$ can be formalised as:
 \begin{align}
 L_{\mathrm{m}}\left(\mathbf{x}_{i}, \mathbf{x}_{j} ; f\right)=\left(1-y_{i j}\right)\left[\alpha-d_{i j}\right]_{+}+y_{i j}\left[d_{i j}-(\alpha-m)\right]_{+}
\end{align}
 where $y_{i j} =1$ when $i$ and $j$ belong to the same class and $y_{i j}=0$ otherwise. The $d_{i j}$ is the Euclidean distance between the embedding of the two points $f(x_i)$ and $f(x_j)$, where $f(x)$ is the embedding function.
 Given a query $x_c$, RLL loss function computes the distance between the query and all other data points followed by ranking these data points according to their similarity to the query. The result of this process is a non-trivial positive list and a non-trivial negative list as the following:
\begin{equation}
   p_{c,i}^*=
   \{
   x_{j}^c |j\neq i, d_{i j} > \alpha - m
   \}
\end{equation} 
\begin{equation}
   N_{c,i}^*=
   \{
   x_{j}^K |k\neq c, d_{i j} < \alpha 
   \}
\end{equation} 
To pull all the non-trivial positive samples close together and to create the hyper-sphere of the class, $L_P$ is minimised and applied as follows:
\begin{equation}
L_{\mathrm{P}}\left(\mathbf{x}_{i}^{c} ; f\right)=\frac{1}{\left|\ P_{c, i}^{*}\right|} \sum_{\mathbf{x}_{j}^{c} \in \mathbf{P}_{c, i}^{*}} L_{\mathrm{m}}\left(\mathbf{x}_{i}^{c}, \mathbf{x}_{j}^{c} ; f\right)
\end{equation}
The non-trivial negative points need to be pushed beyond the boundary $\alpha$ by minimising the following objective function: 
\begin{equation}
L_{\mathrm{N}}\left(\mathbf{x}_{i}^{c} ; f\right)=\sum_{\mathbf{x}_{j}^{k} \in\left|\mathbf{N}_{c, i}^{*}\right|} \frac{w_{i j}}{\sum_{\mathbf{x}_{j}^{k} \in\left|\mathbf{N}_{c, i}^{*}\right|} | w_{i j}} L_{\mathrm{m}}\left(\mathbf{x}_{i}^{c}, \mathbf{x}_{j}^{c} ; f\right)
\end{equation}\\
where $w_{i j}$ is the weight of the negative non-trivial samples. 
To balance between positive and negative objectives RLL is optimized with $\lambda$ as the following:
\begin{equation}
RLL_{\mathrm{L}}\left(\mathbf{x}_{i}^{c} ; f\right)=L_{\mathrm{P}}\left(\mathbf{x}_{i}^{c} ; f\right)+\lambda L_{\mathrm{N}}\left(\mathbf{x}_{i}^{c} ; f\right)
\end{equation}
Using the center loss \cite{wenDiscriminativeFeatureLearning2016a} in Re-ID model is a good practice as it guides the model to learn the centre of deep features for each class and decrease the distances between the embedding and the class they belong to. The center loss can be defined as the following:
\begin{equation}\label{eq:centre_loss}
center_{L}=\frac{1}{2} \sum_{i=1}^B ||f_i-c_{y_i}||_2^2
\end{equation}
where $y_i$ is the label of video $i$ in the mini-batch, while $c_{y_i}$ is the class center of its deep features and $B$ is the batch size. In our case, we have used a weighted center loss to optimise the center loss through all the classes. Using Stochastic Gradient Descent (SGD) optimiser for center loss, the parameters of the classes in the mini-batch are updated in each iteration by averaging the features of these classes. The use of this weighted center loss supports the intra-class distance minimisation by the RLL loss \cite{wangRankedListLoss2019b} with $\beta$ as balanced weight of center loss.   \\
To enhance the impact of Random Erasing Augmentation (REA), a high attention score is given to the frame containing the erased region. Following the methodology proposed by \cite{pathakVideoPersonReID2019a}, by labelling the erased frames by 1 and others by 0 the Erasing-attention loss $E\_att_{L}$ can be calculated as the following:
\begin{equation}
E\_att_{L}= \frac{1}{T} \sum_{t=1}^T E\_Label_t \;\; a_i^t  
\end{equation}\\
where $E\_Label_t$ is $1$ for frame with erased region and $0$ otherwise. The $a_i^t$ is the frame level score given by the temporal attention.
The overall loss function used to optimize our model can be formulated as:
\begin{equation}\label{eq:total_loss}
L= ID_{L} +RLL_{L} +\beta \times center_{L} + E\_att_{L}
\end{equation}
The $ID_{L}$ and $RLL_{L}$ play different roles to guide the model to produce a robust feature representation for person Re-ID. The $ID_{L}$ supports the model to learn a more discriminative  features, while the $RLL_{L}$ is used to make similar samples closer in the embedding space and make dissimilar samples have greater separation using a predefined distance measurement. The $E\_att_{L}$ guides the model to deal with occlusions in the video. Consequently, these loss functions appear with equal weight and different roles in our loss function. On the other hand, the $center_{L}$ is used as support for $RLL_{L}$ to find the center of each class, following the suggestion of \cite{luoStrongBaselineBatch2019} we set the wight of the center loss to $\beta$.
\section{Evaluation}
We evaluate our proposed model on three video-based Re-ID datasets, MARS \cite{zhengMARSVideoBenchmark2016b}, PRID2011 \cite{hirzerPersonReidentificationDescriptive2011b} and	iLIDS-VID \cite{wangPersonReidentificationSystem2014d}. In this section we will introduce the datasets, evaluation metrics, implementation details and the different training choices that affects the performance of our proposed approach.

\subsection{ Datasets and Evaluation Protocols}
Our model is tested on three established benchmark video Re-ID datasets.MARS \cite{zhengMARSVideoBenchmark2016b} is a large-scale dataset that consists of 1261 identities and 20,715 tracklet under 6 cameras and is considered as  the largest video Re-ID benchmark. The bounding boxes are produced by DPM detector \cite{felzenszwalbObjectDetectionDiscriminatively2010} and the GMMCP tracker \cite{dehghanGMMCPTrackerGlobally2015a}. 
The PRID2011 \cite{hirzerPersonReidentificationDescriptive2011b} dataset contains 400 sequences of 200 pedestrians captured by two cameras. The length of the sequence varies from 5 to 675 frames.
In iLIDS-VID \cite{wangPersonReidentificationSystem2014d} dataset, there are 600 image sequences of 300 pedestrians from 2 cameras. Each sequence has variable length ranges from 23 to 192.\newline
The evaluation metrics we use to estimate the performance of our model are: Cumulative Match Characteristic (CMC) and mean Average Precision (mAP). CMC is used to evaluate the capability of the model to find the correct identity within the top-k ranked matches whereby we report rank-1 accuracy for all the datasets. The mAP metric is used to evaluate the performance of the model in multi-shot re-identification datasets such as MARS \cite{zhengMARSVideoBenchmark2016b}.

\subsection{ Implementation Details}
Using ResNet50 \cite{heDeepResidualLearning2016a}, or its variant ResNet50-IBN-a \cite{panTwoOnceEnhancing2018b}, as a video frame features extractor is a good choice in terms of accuracy and performance. We find that ResNet50-IBN-a \cite{panTwoOnceEnhancing2018b} with its ability to maintain effective discriminative features and eliminate appearance variance, which is the most challenging part of the Re-ID process, results in more robust features. Subsequently, we have chosen ResNet50-IBN-a \cite{panTwoOnceEnhancing2018b} pre-trained on ImageNet \cite{he2015delving} as our frame features extractor. Following \cite{sunPartModelsPerson2018d}, the last spatial down-sampling stride is changed to 1, to bring higher spatial resolution without additional parameters and with a low computational cost. As shown in Table \ref{table:1}, ResNet50-IBN-a as frame feature extractor, enhances the accuracy in all datasets compared with ResNet50. 

\setlength{\tabcolsep}{4pt}
\begin{table}[h!]
\begin{center}
\caption{comparing Rank-1 accuracy using ResNet50 \cite{heDeepResidualLearning2016a} and ResNet50-IBN-a \cite{panTwoOnceEnhancing2018b} as a frame feature extractor network.}
\label{table:1}
\begin{tabular}{lll}
\hline\noalign{\smallskip}

    & \begin{tabular}{@{}ll@{}}
                   ResNet50 \cite{heDeepResidualLearning2016a}\\
                       rank-1 \\
             \end{tabular}  

 & \begin{tabular}{@{}ll@{}}
                   ResNet50-IBN-a \cite{panTwoOnceEnhancing2018b}\\
                       rank-1\\
             \end{tabular} \\
\hline

\noalign{\smallskip}

\noalign{\smallskip}

MARS \cite{zhengMARSVideoBenchmark2016b}  &  88.21  &	\textbf{89.02} \\
    PRID2011 \cite{hirzerPersonReidentificationDescriptive2011b} &	94.38 &	\textbf{96.6}\\
iLIDS-VID \cite{wangPersonReidentificationSystem2014d} &	87.33 &	\textbf{89.33}\\
\hline
\end{tabular}
\end{center}
\end{table}
\setlength{\tabcolsep}{1.4pt}

The images of the video frames are resized to $244 \times 112$ and the resized image frame is padded by 10 pixels with zero values. It is then randomly cropped into $244 \times 112$ rectangular images. Each image is flipped horizontally with $0.5$ probability and its RGB channels are normalised by subtracting $(0.485, 0.456, 0.406)$ and dividing by $(0.229, 0.224, 0.225)$, respectively following the standard deviation of ImageNet \cite{he2015delving}.\newline
Our model is trained using four loss functions, ID loss \cite{szegedyRethinkingInceptionArchitecture2016a}, center loss \cite{wenDiscriminativeFeatureLearning2016a}, ranked list loss (RLL) \cite{wangRankedListLoss2019b} and erasing-attention loss \cite{pathakVideoPersonReID2019a}. The ID loss is used as a classification loss to compare prediction of fully connected layer identities and the ground truth labels. The input for the fully connected layer is the video features after normalisation.\newline 
The second loss that we apply to encourage our model to learn is RLL \cite{wangRankedListLoss2019b}, a metric learning loss function. RLL is used to force a distance between negative samples to be greater than specific threshold $\alpha$, in our experiments $\alpha=2.0$. In addition, the positive samples are pulled to be closer than a threshold $\alpha - m$, where $m$ is 1.3 for MARS \cite{zhengMARSVideoBenchmark2016b} and iLIDS-VID \cite{wangPersonReidentificationSystem2014d}, but for PRID2011 \cite{hirzerPersonReidentificationDescriptive2011b} $m$ is 0.04.\newline
 We have also added center loss \cite{wenDiscriminativeFeatureLearning2016a} to our loss with the aim to regularise the distances between the deep features of data points and their corresponding class. To balance its weight we followed the suggestion of \cite{luoStrongBaselineBatch2019} and multiply the center loss by factor $\beta$ =  $0.00005$.\\ 
 Our model training, using the four loss functions is formalised in Equation \ref{eq:total_loss}.
Our model is trained for 120 epochs and is validated every 10 epochs. Adam \cite{kingma2014adam} is used as an optimiser for our model with base learning rate equal to 0.00035 and the warmup strategy is followed as set out in Section \ref{sub:Tr} to change the learning rate during the training phase.

\subsection{Performance Comparison of Different Training Choices}
This section describes the effect of different training choices, as the number of frames in a clip, batch size and transfer learning on the model learning performance.
\subsubsection{Impact of the Clip Length}
The video clip length in the training affects the learning process, so we examine the use of the two-common clip lengths in the literature \cite{gaoRevisitingTemporalModeling2018a},\cite{liMultiScale3DConvolution2019},\cite{subramaniamCoSegmentationInspiredAttention2019a}. For the three datasets we examine the use of 4 and 8 frames as a sequence length of each clip in a video to train the network. Our results as shown in Table \ref{table:2} demonstrate that using 4 frames for each video clip will give the best result at inference stage. As a result, we have used 4 frames for each video clip for the rest of our experiments.

\begin{table*}

   \centering
 
\caption{Comparison against state-of-the-art methods, Memory Usage calculate using \cite{Torchsummary}}
\label{table:5}

\begin{tabular}{llp{2cm}llp{1cm}lp{1cm}llll}
            
\hline\noalign{\smallskip}

	Methods & Publication & 
	
 \begin{tabular}{@{}ll@{}}
 MARS \cite{zhengMARSVideoBenchmark2016b} \\
 rank-1 (mAP)\\
 \end{tabular}
 &
 \begin{tabular}{@{}ll@{}}
  PRID2011 \cite{hirzerPersonReidentificationDescriptive2011b}\\
  rank-1\\
 \end{tabular}
 &
 \begin{tabular}{@{}ll@{}}
 iLIDS-VID \cite{wangPersonReidentificationSystem2014d} \\
  rank-1\\
  
 \end{tabular}
 &
 \begin{tabular}{ll}
     \\
    Input \\
  \end{tabular}
  &
  \begin{tabular}{ll}
    Memory Usage (MB)  \\
    Fore/Backward Pass \\
  \end{tabular}
  &
  \begin{tabular}{ll}
      \\
    Params \\
  \end{tabular}
  &
  \begin{tabular}{ll}
      \\
    Total Size\\
  \end{tabular}

 \\
\hline

\noalign{\smallskip}
SAN \cite{liDiversityRegularizedSpatiotemporal2018a} & CVPR 2018 & 82.3 (65.8)& 93.2 & 80.2 \\
Att-Driven \cite{zhaoAttributedrivenFeatureDisentangling2019} & CVPR 2019	 & 87.0	(78.2) & 93.9 &	86.3 \\
VRSTC \cite{houVRSTCOcclusionFreeVideo2019} & CVPR 2019	 & 88.5 (82.3) & -- & 86.3 	\\
Co-Segment \cite{subramaniamCoSegmentationInspiredAttention2019a}& ECCV 2019 &	84.9 (79.9) & --& -- \\
GLTR \cite{liGlobalLocalTemporalRepresentations2019a}& ICCV 2019&	87.02 (78.47) & 95.50 &86.00&9.00&214.11& 94.47&317.59\\
M3D \cite{liMultiScale3DConvolution2019}& 	IEEE-T IP 2020 & 88.63 (79.46) &96.60&86.67&9.00&1213.83 &104.58 &1327.41\\

VPRFT \cite{pathakVideoPersonReID2019a}& AAAI 2020 &	88.6(82.9) & 93.3&-- &9.19 &  153.92 & 290.58 & 453.69\\
Ours &--&\textbf{89.62 (84.61)}  &96.6& 89.33 & 9.19 &  171.92& 290.58 & 471.69 \\
\hline
VPRFT \cite{pathakVideoPersonReID2019a} \\(pre-trained on MARS) & AAAI 2020&	-- & 96.6 &-- \\
Ours \\(pre-trained MARS) &--&--&96.63 &	\textbf{97.33}\\
Ours \\(pre-trained MARS and iLIDS-VID)&-- & 88.21(83.10) & \textbf{97.75}& 95.33\\
\hline
\end{tabular}

\end{table*}

\setlength{\tabcolsep}{4pt}
\begin{table}[H]
\begin{center}
\caption{Performance of Re-ID model with different clip length}
\label{table:2}
\begin{tabular}{llp{2cm}llp{2cm}}
\hline\noalign{\smallskip}
 
 &
 \begin{tabular}{@{}ll@{}}
 MARS \cite{zhengMARSVideoBenchmark2016b} \\
 rank-1 (mAP)\\
 \end{tabular}
 &
 \begin{tabular}{@{}ll@{}}
  PRID2011 \cite{hirzerPersonReidentificationDescriptive2011b}\\
  rank-1\\
 \end{tabular}
 &
 \begin{tabular}{@{}ll@{}}
 iLIDS-VID \cite{wangPersonReidentificationSystem2014d} \\
  rank-1\\
 \end{tabular}
 \\
\hline

\noalign{\smallskip}

\noalign{\smallskip}
4 frames &	\textbf{89.02 (83.9)} &	\textbf{96.6} &	\textbf{89.33} \\
8 frames & 88.70 (82.71) &	95.51&	88.00\\

\hline
\end{tabular}
\end{center}
\end{table}

\subsubsection{Impact of Batch Size}
We use a random sampler in each epoch to sample the mini batch to be $B= C \times K$, where $C$ is the number of classes (person subjects) and $K$ is the number of videos for each class. We studied the influence of the batch size and the number of samples on the Re-ID model accuracy. We have examined the use of different sizes with different structures of the batch. Starting from the most common batch size for video Re-ID, 32, as batch size with 8 classes and 4 instances from each class followed by increasing the batch size to 36 and 48 with $C=6$ and $K=$ 6 and 8 respectively. Overall, we have intensely experimented our model using different batch sizes by examining the effect of the total batch size 32, 36, 48. The results are shown in Table \ref{table:3} where it can be seen that the batch size does not play a crucial role at our proposed Re-ID method as the differentiation in Rank-1 accuracy is less than $1\%$ in all datasets. As a result, our final results use a batch size of 32 as to allow a fair comparison with other studies that use 32 as batch size \cite{pathakVideoPersonReID2019a}, \cite{zhaoAttributedrivenFeatureDisentangling2019}, \cite{houVRSTCOcclusionFreeVideo2019}, \cite{subramaniamCoSegmentationInspiredAttention2019a}. 

\setlength{\tabcolsep}{6pt}
\begin{table}[H]
\begin{center}

\caption{Comparison rank-1 performance with differing batch size}
\label{table:3}
\resizebox{9cm}{!}{
\begin{tabular}{llll}
\hline\noalign{\smallskip}
  &
 \begin{tabular}{@{}ll@{}}
 MARS \cite{zhengMARSVideoBenchmark2016b} \\
 rank-1 (mAP)\\
 \end{tabular}
 &
 \begin{tabular}{@{}ll@{}}
  PRID2011 \cite{hirzerPersonReidentificationDescriptive2011b}\\
  rank-1\\
 \end{tabular}
 &
 \begin{tabular}{@{}ll@{}}
 iLIDS-VID \cite{wangPersonReidentificationSystem2014d} \\
  rank-1\\
 \end{tabular}
 \\
    
\hline

\noalign{\smallskip}

\noalign{\smallskip}
Batch size 32, C= 8, K=4 instances &	89.02 (83.9) &	\textbf{96.6} &	89.33 \\
Batch size 36, C= 6, K=6 instances &	\textbf{89.62} (84.61) &	93.26 &	\textbf{90.00}\\
Batch size 48, C= 6, K=8 instances&	89.35 \textbf{(84.71)} &  95.51 &	89.33\\
\hline
\end{tabular}
}
\end{center}
\end{table}

\subsubsection{Impact of Transfer Learning} \label{sec:ev}
In order to assess the impact of transfer learning we perform pre-training on both ImageNet \cite{he2015delving} and each of the datasets considered with results shown in 
Table \ref{table:4}. The results show significant improvement on iLIDS-VID \cite{wangPersonReidentificationSystem2014d} when the model pre-trained on MARS \cite{zhengMARSVideoBenchmark2016b}, whereas on the other hand there is only slight improvement on PRID2011 \cite{hirzerPersonReidentificationDescriptive2011b}. We have also examined fine-tuning our model using a sequence of pre-training on iLIDS-VID \cite{wangPersonReidentificationSystem2014d} and MARS \cite{zhengMARSVideoBenchmark2016b}. These results register an improvement on PRID2011 \cite{hirzerPersonReidentificationDescriptive2011b}, as shown in 
Table \ref{table:4}. However, MARS \cite{zhengMARSVideoBenchmark2016b}  did not benefit from the pre-trained model on other Re-ID datasets, we attribute this to the larger size of MARS \cite{zhengMARSVideoBenchmark2016b} and the high level of challenge it represents.

\setlength{\tabcolsep}{1.4pt}
\setlength{\tabcolsep}{4pt}
\begin{table}[H]
\begin{center}
\caption{Comparison rank-1 value on MARS \cite{zhengMARSVideoBenchmark2016b}, PRID2011 \cite{hirzerPersonReidentificationDescriptive2011b} and iLIDS-VID \cite{wangPersonReidentificationSystem2014d} using transfer learning}
\label{table:4}
\resizebox{9cm}{!}{
\begin{tabular}{llll}
\hline\noalign{\smallskip}

	  &
 \begin{tabular}{@{}ll@{}}
 MARS \cite{zhengMARSVideoBenchmark2016b} \\
 rank-1 (mAP)\\
 \end{tabular}
 &
 \begin{tabular}{@{}ll@{}}
  PRID2011 \cite{hirzerPersonReidentificationDescriptive2011b}\\
  rank-1\\
 \end{tabular}
 &
 \begin{tabular}{@{}ll@{}}
 iLIDS-VID \cite{wangPersonReidentificationSystem2014d} \\
  rank-1\\
 \end{tabular}
 \\
    
\hline
\noalign{\smallskip}
\noalign{\smallskip}
pre-trained ImageNet \cite{he2015delving} & \textbf{89.02 (83.9)}  & 96.6 & 89.33 \\
pre-trained MARS \cite{zhengMARSVideoBenchmark2016b} &   --  &	 96.63 & \textbf{97.33}	\\
pre-trained on MARS \cite{zhengMARSVideoBenchmark2016b} + ilIDS-VID \cite{wangPersonReidentificationSystem2014d} &  88.21(83.10) & \textbf{97.75} & 95.33 \\ 
\hline
\end{tabular}
}
\end{center}

\end{table}

\subsection{Results}
Table \ref{table:5} reports the performance of our approach and other state-of-the-art methods on MARS \cite{zhengMARSVideoBenchmark2016b}, PRID2011 \cite{hirzerPersonReidentificationDescriptive2011b} and iLIDS-VID \cite{wangPersonReidentificationSystem2014d}, respectively. On all the three datasets our approach outperforms the best existing state-of-the-art methods \cite{zhaoAttributedrivenFeatureDisentangling2019},\cite{houVRSTCOcclusionFreeVideo2019},\cite{subramaniamCoSegmentationInspiredAttention2019a},\cite{liMultiScale3DConvolution2019},\cite{pathakVideoPersonReID2019a}. We attribute improved performance to the combined use of best practices strategies such as BNNeck \cite{luoStrongBaselineBatch2019}, Random Erasing Augmentation (REA) \cite{zhongRandomErasingData2017a}, Warmup \cite{fanSphereReIDDeepHypersphere2019a}, in addition to transfer learning with our simple single stream model that uses only global features from each image frame and attention based clip level aggregation. This simple single stream architecture, combined with the use of these best practice strategies, has been shown to outperform the 3D convolution based approaches of \cite{liMultiScale3DConvolution2019} and the complex multiple stream architectures of \cite{chenVideoPersonReidentification2018b},\cite{mclaughlinRecurrentConvolutionalNetwork2016b},\cite{xuJointlyAttentiveSpatialtemporal2017}, within a lower computational complexity and memory footprint.\\
\begin{figure}
  \renewcommand{\figurename}{Figure}
  \includegraphics[width=\linewidth, height=12cm]{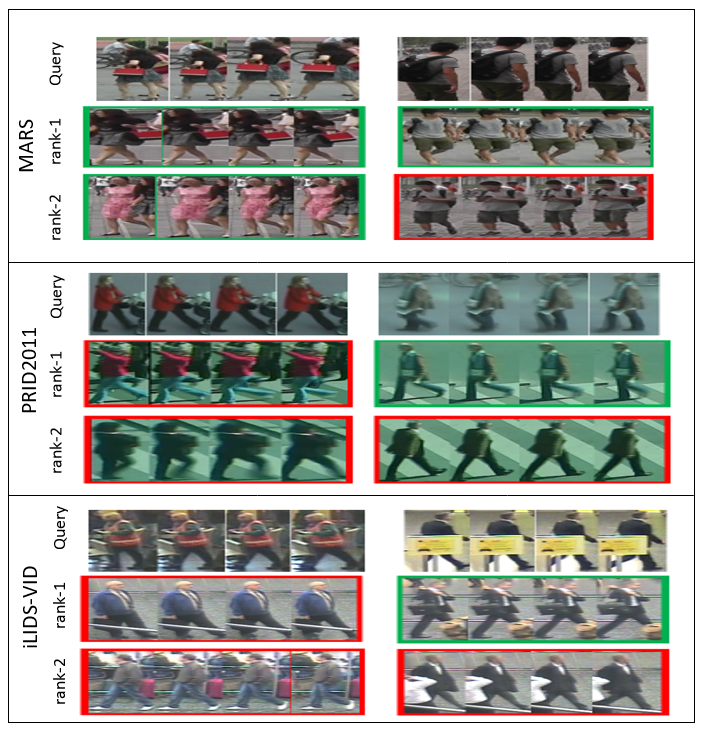}
  \caption{Illustrative examples of rank-1 and rank-2 given a query (Green is true match and red is false match)}
  \label{Figure 2}
\end{figure}

\section{Conclusions}
In this paper, we propose a single stream robust video Re-ID approach based on the use of 2D convolution using only global features with simple temporal attention that is trained using multiple effective loss functions, which are Label Smoothing (LS) loss \cite{szegedyRethinkingInceptionArchitecture2016a}, RLL loss \cite{wangRankedListLoss2019b}, center loss \cite{wenDiscriminativeFeatureLearning2016a}, and erasing-attention loss \cite{pathakVideoPersonReID2019a}, resulting in a strong video level feature representation for video Re-ID task. 
Effective training strategies are also applied to direct our model to learn an efficient video feature representation. We employ, BNNeck \cite{luoStrongBaselineBatch2019}, Random Erasing Augmentation(REA) \cite{zhongRandomErasingData2017a}, Warmup \cite{fanSphereReIDDeepHypersphere2019a} and transfer learning.\\
Our model achieves state-of-the-art performance with $89.62\%$, $97.75\%$, $97.33\%$ rank-1 accuracy on MARS \cite{zhengMARSVideoBenchmark2016b}, PRID2011 \cite{hirzerPersonReidentificationDescriptive2011b} and iLIDS-VID \cite{wangPersonReidentificationSystem2014d} datasets respectively, with $84.61\%$ mAP for MARS \cite{zhengMARSVideoBenchmark2016b} and outperforms the prior work of \cite{liDiversityRegularizedSpatiotemporal2018a},\cite{zhaoAttributedrivenFeatureDisentangling2019},\cite{houVRSTCOcclusionFreeVideo2019},\cite{subramaniamCoSegmentationInspiredAttention2019a},\cite{liGlobalLocalTemporalRepresentations2019a},\cite{liMultiScale3DConvolution2019},\cite{pathakVideoPersonReID2019a},\cite{pathakVideoPersonReID2019a}, notably the prior 3D convolution based works of \cite{liMultiScale3DConvolution2019} and the complex multi-stream works of \cite{chenVideoPersonReidentification2018b},\cite{mclaughlinRecurrentConvolutionalNetwork2016b},\cite{xuJointlyAttentiveSpatialtemporal2017}. \\
In future work, we will continue to simplify video based person Re-ID methods, targeting the requirement of real-time systems. Our main focus is to use simple backbone network architectures and increase the use of robust training strategies to improve performance. In addition, as person Re-ID can be considered largely solved for some of the datasets as PRID2011 \cite{hirzerPersonReidentificationDescriptive2011b} and iLIDS-VID \cite{wangPersonReidentificationSystem2014d}, with $97.75\%$, $97.33\%$ rank-1 accuracy respectively, we will investigate our method in the challenging airport \cite{gou2018systematic} dataset with small video frames of size $128 \times 64$.

\bibliographystyle{IEEEtran}
\bibliography{paper}
\end{document}